\documentclass[runningheads]{llncs}
\usepackage{graphicx}
\begin{document}
\title{Better AI through Logical Scaffolding}
%
%
\author{Nikos Ar\'echiga\inst{1}
\and Jonathan DeCastro\inst{1}
\and Soonho Kong\inst{1}
\and Karen Leung\inst{2}
}
\authorrunning{N. Ar\'echiga et al.}
%
\institute{Toyota Research Institute, Los Altos, CA and Cambridge, MA, USA \and
Stanford University, Palo Alto, CA, USA
}
\maketitle              
\begin{abstract}
We describe the concept of \emph{logical scaffolds}, which can be used to improve the quality of software that relies on AI components. We explain how some of the existing ideas on runtime monitors for perception systems can be seen as a specific instance of logical scaffolds. Furthermore, we describe how logical scaffolds may be useful for improving AI programs beyond perception systems, to include general prediction systems and agent behavior models.
\end{abstract}

\section{Introduction}

Recent progress in AI has led to possible deployment in a wide variety
of important domains. This includes safety-critical cyberphysical
systems such as automobiles~\cite{bojarski_end_2016} and
airplanes~\cite{julian_policy_2016}, but also decision making systems in diverse domains including
legal~\cite{perry_predictive_2013} and military
applications~\cite{cummings_artificial_2017}.

Current AI programs differ from traditional programs in their reliance
on data. The specification, input-output semantics, and executable
generation procedure are all data driven~\cite{karpathy_software_2017}.

Unlike traditional software development, in AI programs a specification
is not formally articulated. Indeed, in many of the most promising
recent applications of AI, such as vision and human intent prediction,
it is not feasible to write a formal specification. Instead, an implicit
specification is provided via a test set, and the goal is to achieve a
certain performance over the test set.

Traditional software development specifies the input-output semantics of
the program in a programming language. In AI programs, the engineer
provides a training dataset, and the program must match the input-output
statistics of the dataset.

Instead of using a compiler to translate the programming language
constructs to machine code, an engineer provides a ``skeleton'' in the
form of a powerful function approximator (such as a neural network). The
engineer then uses an optimization procedure to search for the
parameterization that provides the best approximation to the input-output
statistics of the training data. A cross-validation set is used to check
generalizability of the learned function to unseen data.

This paradigm has proven to be powerful, especially in domains in which
it is difficult to formally articulate the specification for the
program, much less write a declarative program. However, this approach
suffers from the following drawbacks.

\begin{enumerate}
\item {\bf Implicit specifications} Since the specification is given
implicitly as a desired performance over a test set, it is difficult for
the tests to ascertain whether the program is providing the right
answers for the right reasons~\cite{koopman_toward_2018}. For this
reason, the test set may fail to test the right things. This type of
implicit specification is inadequate for use in a safety case, and it
will be difficult to ascertain that programs tested in this way will be
safe to deploy.

\item {\bf Non-representative training set} Since the program seeks to
match the statistical input-output properties of the training dataset,
deficiencies of this training set will extend to deficiencies of the
program. For example, scenarios that occur rarely in the training set
may be fairly common in the real world, leading to degraded performance
in deployment~\cite{cui_class-balanced_2019}. In this sense, the
training set may fail to train for realistic scenarios.

\item {\bf Robustness and sensitivity to adversarial attacks} Since the
model is templated by a
functional template with many degrees of freedom, it is common for the
process to result in programs that are susceptible to extreme
sensitivity to irrelevant features of the input. The literature on
\emph{adversarial examples} demonstrates how slight perturbations to an
input can lead to incorrect results with high
confidence~\cite{goodfellow_explaining_2015}.

\end{enumerate}

We propose to attack these issues by the use of \emph{logical
scaffolds}, which are lightweight formal properties that
provide some information about the relationship of the program inputs
and outputs. These logical scaffolds can be written in languages for
which monitoring algorithms exist, such as Signal Temporal
Logic~\cite{maler_monitoring_2004}, Signal Convolutional
Logic~\cite{silvetti_signal_2018}, Timed Quality Temporal
Logic~\cite{dokhanchi_evaluating_2018} and many others. Logical
scaffolds may arise from a number of different sources, including a
formalization of physical laws, domain knowledge, and common sense.

Logical scaffolds are more general than related work such as reasonableness
monitors~\cite{gilpin_reasonableness_2018} and model
assertions~\cite{kang_model_2018} because scaffolds can be used for
different types of AI programs beyond merely perception.  Furthermore,
recent work in smoothly differentiable formulations of STL and MTL~\cite{leung_backpropagation_2019,pant_smooth_2017,mehdipour_arithmetic-geometric_2019} enable the scaffolds to become part
of the training process directly.

\section{Logical scaffolds}

Informally, a logical scaffold is a predicate that encodes something
that is believed to be true about the input-output relation of an AI
program. It is \emph{not} a complete specification. If a specification
existed, the scaffold would be a logical consequence of the complete
specification. In other words, the scaffold is a consequence of correct
functionality. As such, it constitutes a necessary, but not sufficient,
condition for correct behavior.

Most of the existing literature on monitoring runtime properties is
centered around perception systems. These runtime monitors can be
formalized as scaffolds, but the key idea can be generalized beyond
perception to include applications in explainable intent prediction and
expressible behavior modeling.

Sources of logical scaffolds are as diverse as the sources of human
intuition about the application domain. The following is not an
exhaustive list.

\begin{itemize}
\item \textbf{Perception}
\begin{itemize}
\item Commonsense notions of label consistency, like the properties
monitored in~\cite{dokhanchi_evaluating_2018} and~\cite{kang_model_2018}, in which class labels are not expected to
mutate or drop out between frames.

\item Class-specific information, such as the intuition that a mailbox
should not be seen crossing the road~\cite{gilpin_reasonableness_2018}.
\end{itemize}

\item \textbf{Intent prediction and behavior modeling}
\begin{itemize}
\item Physics-derived knowledge, such as knowledge of maximum speeds or
actuator capabilities, for example, that on an icy road, other vehicles may
be out of control or less able to brake and swerve.

\item Natural expectations that pedestrians and vehicles are unlikely to
seek damage to themselves, unless they are out of control.
\end{itemize}
\end{itemize}

The challenges related to implicit specifications are
ameliorated by using logical scaffolds at training and testing time.
For generative models, such as intent predictors and agent models, we
can impose an understandable structure on the latent space, as described
below. At testing time, we are able to use parametric logical scaffolds
to learn explanations of the input-output behavior of the system, which
can be used to check that the system is passing its tests for the right
reasons.

The challenges related to non-representative training sets as well as
robustness and sensitivity to adversarial attacks can be ameliorated by
using the scaffolds at deployment.

The work of Kang et al.~\cite{kang_model_2018} has shown how hand crafted
runtime monitors can be used to flag scenarios in which the program
fails. These scenarios can then be added to the training set, yielding
greatly improved performance. This approach has a flavor of active
learning, in which the scenarios that are most difficult for the program
are fed back for further study. Here, we propose that the monitors need
not be hand crafted, but automatically generated from logical scaffolds
that express a variety of properties.  Conversely,
in~\cite{dokhanchi_evaluating_2018}, Dokhanchi et al. automatically
generate monitors from Timed Quality Temporal Logic that check for
label stability, i.e., ensuring that labels do not mutate across frames.

\subsection{Training}
There may be many ways that logical scaffolds can be used at training
time. In this work we consider training generative models that make use
of a latent space.

Generative models are models that can generate data that is similar to
the data they are trained on. Important examples of generative models
for contemporary applications include the following.

\begin{enumerate}
\item Intent predictors, which are used by autonomous vehicles or other
robots to predict future trajectories of other agents, such as
automobiles, pedestrians and bicycles.  

\item Reactive agents, which seek to generate appropriate behaviors for
a specific environment. Examples may include simulations agents that
subject an autonomous vehicle, drone, or robot to challenging but
realistic behaviors, as well as decision-making agents for different
applications.

\item Scenario generators, which seek to synthesize testing and
simulation environments that may be challenging but still realistic.
\end{enumerate}

\begin{figure}[ht]
	\centering
	\includegraphics[width=\textwidth,trim={0cm 17cm 31cm 12cm},clip]{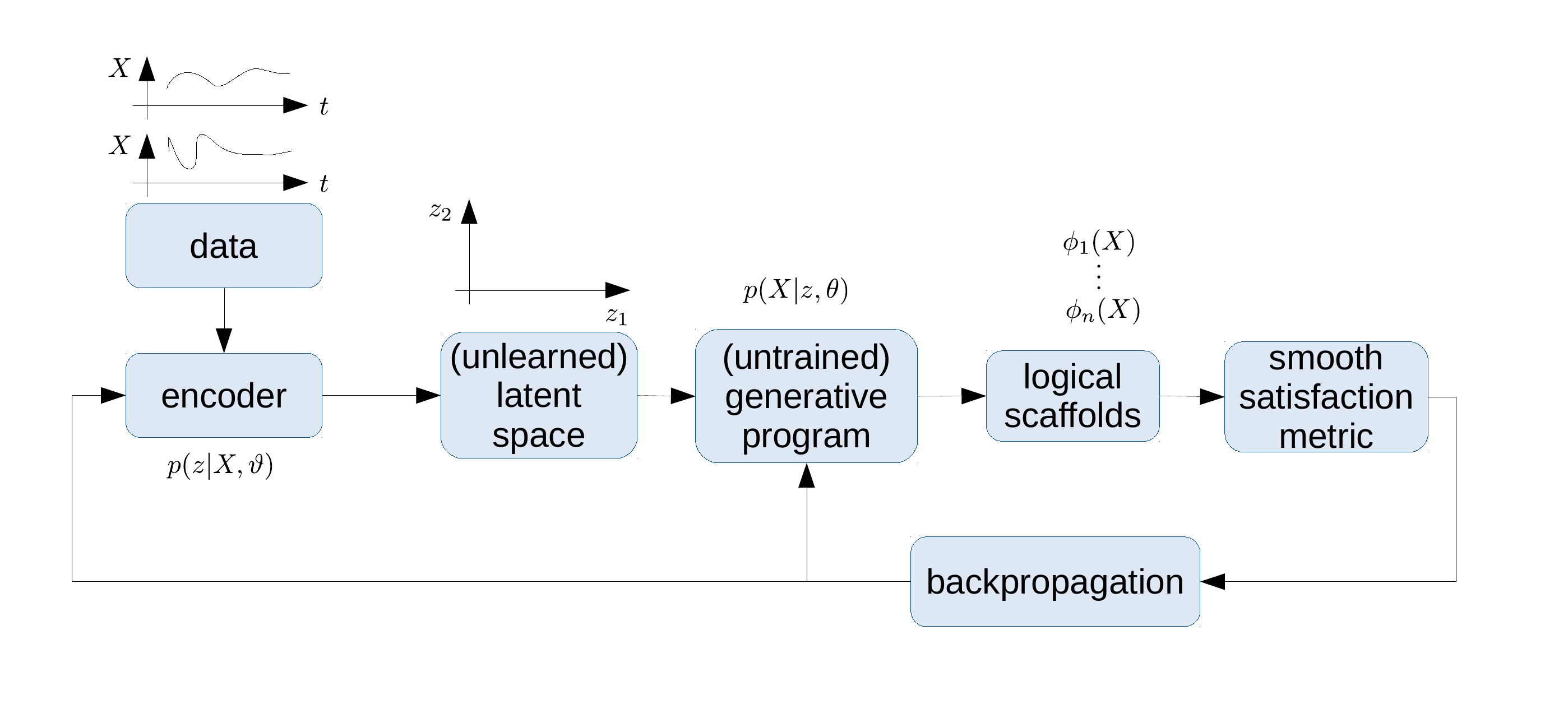}
	\caption{Learning latent spaces from explanations}
	\label{x2latent}
\end{figure}

We can use logical scaffolds expressed in a differentiable logic (e.g. \cite{leung_backpropagation_2019}, \cite{mehdipour_arithmetic-geometric_2019}, \cite{pant_smooth_2017}) and use them to add structure to the latent
space. A diagram of this idea is shown in Figure~\ref{x2latent}.


\subsection{Testing}
At test time, we are interested in finding out whether the model is
producing the correct outputs for the correct reasons. To accomplish
this, we make use of \emph{parametric} logical scaffolds, i.e. scaffolds
with free parameters.

\begin{figure}[ht]
	\centering
	\includegraphics[width=\textwidth,trim={0cm 31cm 32cm 0cm},clip]{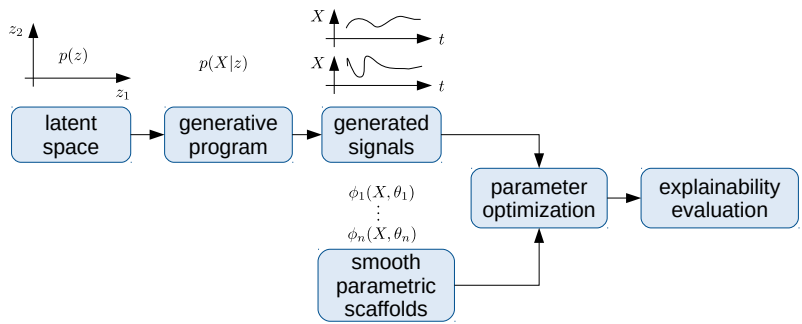}
	\caption{Learning explanations for latent spaces}
	\label{latent2x}
\end{figure}

We sample from the latent space, and prompt the model to generate an
output. Then, we take a bank of pSTL formulas, and fit values for each
of the parameters. Then, we check to see which of the parameters have
clusters that correspond to the clusters of the original latent space.
The corresponding STL formulas are ``explanations'' of the latent space
clusters.  A diagram of this idea is shown in Figure~\ref{latent2x}.

\subsection{Deployment}

At deployment, the logical scaffold can be used to detect anomalies.
These anomalies can later be fed back into the training procedure for
the next iteration of the system.  The use of logical scaffolds for
runtime improvement is predicated on the fact that these scaffolds are
written in formal languages that support runtime monitoring.

The work of~\cite{kang_model_2018} has already demonstrated how
hand-crafted runtime monitors can be used to greatly improve the
performance of single shot detectors, and the work
of~\cite{dokhanchi_evaluating_2018} has developed a special-purpose
runtime logic to monitor the stability of class labels.

The idea of augmenting AI programs with knowledge of physics (``Newton +
Hinton''\footnote{Thanks to Adrien Gaidon for this term}) is not new. We
believe that even greater impact can be obtained from the broader
principle of systematically developing logical scaffolds that encode
physics domain-specific knowledge, and common sense.

\section{Conclusions and Future Work}
We have outlined a technique of ``logical scaffolding'', which involves conditions that are necessary for correctness, but not sufficient. We have outlined how these logical scaffolds can be used to improve the performance and reliability of AI systems at training, testing, and deployment.

In future work, we will explore case studies in perception, behavior modeling, and scenario generation.

\bibliographystyle{plain}
\bibliography{FoMLaS19}

\begin{thebibliography}{10}

\bibitem{bojarski_end_2016}
Mariusz Bojarski, Davide Del~Testa, Daniel Dworakowski, Bernhard Firner, Beat
  Flepp, Prasoon Goyal, Lawrence~D. Jackel, Mathew Monfort, Urs Muller, Jiakai
  Zhang, Xin Zhang, Jake Zhao, and Karol Zieba.
\newblock End to {End} {Learning} for {Self}-{Driving} {Cars}.
\newblock {\em arXiv:1604.07316 [cs]}, April 2016.
\newblock arXiv: 1604.07316.

\bibitem{cui_class-balanced_2019}
Yin Cui, Menglin Jia, Tsung-Yi Lin, Yang Song, and Serge Belongie.
\newblock Class-{Balanced} {Loss} {Based} on {Effective} {Number} of {Samples}.
\newblock {\em arXiv:1901.05555 [cs]}, January 2019.
\newblock arXiv: 1901.05555.

\bibitem{cummings_artificial_2017}
M~L Cummings.
\newblock Artificial {Intelligence} and the {Future} of {Warfare}.
\newblock Technical report, Chatham House, 2017.

\bibitem{dokhanchi_evaluating_2018}
Adel Dokhanchi, Heni~Ben Amor, Jyotirmoy~V. Deshmukh, and Georgios Fainekos.
\newblock Evaluating {Perception} {Systems} for {Autonomous} {Vehicles} {Using}
  {Quality} {Temporal} {Logic}.
\newblock In {\em Runtime {Verification}}, volume 11237. Springer International
  Publishing, Cham, 2018.

\bibitem{gilpin_reasonableness_2018}
Leilani~H. Gilpin.
\newblock Reasonableness {Monitors}.
\newblock {\em AAAI}, 2018.

\bibitem{goodfellow_explaining_2015}
Ian~J. Goodfellow, Jonathon Shlens, and Christian Szegedy.
\newblock Explaining and {Harnessing} {Adversarial} {Examples}.
\newblock In {\em {arXiv}:1412.6572 [cs, stat]}, 2015.
\newblock arXiv: 1412.6572.

\bibitem{julian_policy_2016}
Kyle~D. Julian, Jessica Lopez, Jeffrey~S. Brush, Michael~P. Owen, and Mykel~J.
  Kochenderfer.
\newblock Policy compression for aircraft collision avoidance systems.
\newblock In {\em 2016 {IEEE}/{AIAA} 35th {Digital} {Avionics} {Systems}
  {Conference} ({DASC})}. IEEE, September 2016.

\bibitem{kang_model_2018}
Daniel Kang, Deepti Raghavan, Peter Bailis, and Matei Zaharia.
\newblock Model {Assertions} for {Debugging} {Machine} {Learning}.
\newblock {\em NeurIPS MLSys Workshop}, 2018.

\bibitem{karpathy_software_2017}
Andrej Karpathy.
\newblock Software 2.0, November 2017.

\bibitem{koopman_toward_2018}
Philip Koopman and Michael Wagner.
\newblock Toward a {Framework} for {Highly} {Automated} {Vehicle} {Safety}
  {Validation}.
\newblock In {\em {SAE} {World} {Congress}}, April 2018.

\bibitem{leung_backpropagation_2019}
Karen Leung, Nikos Arechiga, and Marco Pavone.
\newblock Backpropagation for {Parametric} {STL}.
\newblock In {\em Intelligent {Vehicles} {Symposium}}, 2019.

\bibitem{maler_monitoring_2004}
Oded Maler and Dejan Nickovic.
\newblock Monitoring {Temporal} {Properties} of {Continuous} {Signals}.
\newblock In {\em Formal {Techniques}, {Modelling} and {Analysis} of {Timed}
  and {Fault}-{Tolerant} {Systems}}, 2004.

\bibitem{mehdipour_arithmetic-geometric_2019}
Noushin Mehdipour, Cristian~Ioan Vasile, and Calin Belta.
\newblock Arithmetic-geometric mean robustness for control from signal temporal
  logic specifications.
\newblock {\em CoRR}, abs/1903.05186, 2019.

\bibitem{pant_smooth_2017}
Yash~Vardhan Pant, Houssam Abbas, and Rahul Mangharam.
\newblock Smooth operator: {Control} using the smooth robustness of temporal
  logic.
\newblock In {\em 2017 {IEEE} {Conference} on {Control} {Technology} and
  {Applications} ({CCTA})}, pages 1235--1240, Mauna Lani Resort, HI, USA,
  August 2017. IEEE.

\bibitem{perry_predictive_2013}
Walter~L. Perry, Brian McInnis, Carter~C. Price, Susan~C. Smith, and John~S.
  Hollywood.
\newblock {\em Predictive policing: the role of crime forecasting in law
  enforcement operations}.
\newblock RAND Corporation, Santa Monica, CA, 2013.

\bibitem{silvetti_signal_2018}
Simone Silvetti, Laura Nenzi, Ezio Bartocci, and Luca Bortolussi.
\newblock Signal {Convolution} {Logic}.
\newblock {\em ATVA}, 2018.
\newblock arXiv: 1806.00238.

\end{thebibliography}

\end{document}